\documentclass[compactaffiliation]{Interspeech}

\usepackage{subfigure}

\interspeechcameraready

\title{FROST-EMA: Finnish and Russian Oral Speech Dataset of Electromagnetic Articulography Measurements with L1, L2 and Imitated L2 Accents}

\author[affiliation={1}]{Satu}{Hopponen}
\author[affiliation={2}]{Tomi}{Kinnunen}
\author[affiliation={1}]{Alexandre}{Nikolaev}
\author[affiliation={1,3}]{Rosa}{González Hautamäki}
\author[affiliation={4}]{Lauri}{Tavi}
\author[affiliation={5}]{Einar}{Meister}

\affiliation{School of Humanities}{University of Eastern Finland}{Finland}
\affiliation{School of Computing}{University of Eastern Finland}{Finland}
\affiliation{Faculty of Humanities}{University of Oulu}{Finland}
\affiliation{}{Forensic Laboratory of National Bureau of Investigation}{Finland}
\affiliation{}{Tallinn University of Technology}{Estonia}
\email{shoppone@student.uef.fi, tomi.kinnunen@uef.fi, alexandre.nikolaev@uef.fi, Rosa.GonzalezHautamaki@oulu.fi, lauri.tavi@poliisi.fi, einar.meister@taltech.ee}

\usepackage{comment,wasysym}

\begin{document}

\maketitle

\begin{abstract}
We introduce a new FROST-EMA (Finnish and Russian Oral Speech Dataset of Electromagnetic Articulography) corpus. It consists of 18 bilingual speakers, who produced speech in their native language (L1), second language (L2), and imitated L2 (fake foreign accent). The new corpus enables research into language variability from phonetic and technological points of view. Accordingly, we include two preliminary case studies to demonstrate both perspectives. The first case study explores the impact of L2 and imitated L2 on the performance of an automatic speaker verification system, while the second illustrates the articulatory patterns of one speaker in L1, L2, and a fake accent.
\end{abstract}

\section{Introduction} 

Humans produce speech by coordinating the movements of various \emph{articulators}, such as the tongue and the lips. Humans continuously reshape their vocal tract to modulate the airstream produced by the voice source, the larynx. The resulting acoustic waveform, commonly captured with a microphone, contains latent (i.e. not directly observable) information of articulatory kinematics. 
A number of alternative techniques to track articulatory movements are available, including x-ray \cite{badin_1991}, ultrasound \cite{Nance_2024}, magnetic resonance imaging \cite{Narayanan2014-realtime}, and \emph{electromagnetic articulography} (EMA). In this study, we focus on EMA, which is a 3D measurement technique that tracks the movements of sensor coils attached to the articulators with an adhesive. 

Detailed analysis of articulation, which EMA facilitates, is relevant to the study of speech variability. The human vocal tract is highly flexible and can accommodate a wide variation in languages, accents, and speaking styles, to name but a few. 
One interesting phenomenon is \emph{foreign accent}. Second language (L2) learners often apply phonemic substitution patterns from their first language (L1), resulting in speech that blends characteristics of both L1 and L2 \cite{fan2014impact}. Analyzing the articulatory mechanisms behind L1 and L2 speech production can help identify challenges faced by second language learners.

Sometimes \emph{fake} foreign accents are also produced on purpose in movies, stand-up comedy, and social media. From a linguistic point of view, understanding the production of successfully imitated L2 accents can provide novel insights into accent stereotypes. 

The new corpus that we introduce specifically addresses \textbf{variability in the first language (L1), second language (L2) and imitated L2} (fake foreign accent) in \textbf{bilingual speakers}, who are Finnish and Russian speakers with varying degrees of L2 proficiency. The phonetic premises---including the reasons for the specific language choices---are provided in Section \ref{sec:phonetical-basis-for-our-data}.

Importantly, EMA is not relevant only to phonetic sciences but for the development of novel educational and clinical speech applications. In particular, parallel acoustic and articulatory recordings 
enable \emph{articulatory inversion} \cite{Shahrebabaki2022-Articulatory-Inversion}, which is the task of estimating articulation movements from the acoustic data, and \emph{articulatory synthesis} \cite{birkholz_2013}, which is the task of generating speech based on articulatory parameters. 
Articulation can also help \cite{tavi_2021} to shine a light on the weak points of automatic speaker recognition systems, when faced with purposefully disguised voices \cite{neuhauser2008voice}. This is a pressing issue in forensic voice comparison. Another promising 
use of EMA is its potential in \emph{prior} kinematic speech knowledge in specific models, which can be utilized in relaxing the acoustic training data requirements. 
For instance, \cite{Anumanchipalli2019-neural-decoding} utilized EMA data as part in a brain-to-speech synthesis to address data limitations. 
Another interesting study \cite{Cho2023-Articulation-SSL} used \emph{linear probing} to study the association of EMA coordinate values with the latent features extracted from various foundational speech models. Their results had remarkably high Pearson correlations. 
Some foundational models appear to encode 
articulatory information even if they were not trained to do so. Recent articulatory inversion approaches 
already use foundational speech models to generate robust features for inversion tasks. The above studies demonstrate the potential role of articulation as a bridge between speech science and technology.

Despite its relevance to both fundamental and applied speech research, the currently available EMA resources are surprisingly scarce. Most of them (Table \ref{tab:prior-EMA-databases}) are in English and contain only a handful of subjects. To foster research in both speech science and technology, we introduce a new \textbf{FROST-EMA} database (\emph{Finnish and Russian Oral Speech Dataset of Electromagnetic Articulography Measurements}). The novelties, compared to prior publicly available EMA corpora, are:
(1) the two languages (Finnish and Russian) are not included in existing EMA databases\footnote{Apart from our early pilot \cite{tavi_2021}, which contained only 4 speakers and was collected with an older EMA device.}; (2) the inclusion of \emph{bilingual} speakers who all produce L1, L2, and imitated L2 accents; (3) an overall larger number of subjects compared to existing corpora; (4) the use of a state-of-the-art EMA device with a higher sampling rate. The first two very particular constraints, combined with the practical difficulties in recruiting volunteers for invasive and time-consuming EMA recordings, makes the compilation a corpus a tedious task.

The study has been reviewed and approved by the Ethics Board of the authors' institute. While this Interspeech 2025 paper provides a snapshot of an ongoing data collection effort, our plan is to release the FROST-EMA corpus to the community by the first half of 2026 at the latest.

\section{The collection of FROST-EMA dataset}

\subsection{Phonetic basis for the collected data}\label{sec:phonetical-basis-for-our-data}

Finnish and Russian, though geographically adjacent, belong to different language families—Finno-Ugric and Indo-European \cite{georg_2023}, respectively—and exhibit distinct phonetic and phonological characteristics (for Finnish, see \cite{grunthal_2023} and for Russian, see \cite{Sussex_2006}). Finnish features a larger vowel inventory with a long-short vowel distinction, whereas Russian has an average vowel inventory without such a contrast. In terms of consonants, Russian possesses a broader set of phonemes, while Finnish distinguishes between single and geminate consonants. Additionally, Finnish employs a fixed stress pattern, whereas Russian has variable stress. These contrasting phonetic properties provide a rich framework for investigating both natural and imitated accents.


\renewcommand{\arraystretch}{1.0}
\begin{table*}[tb]
  \caption{Available EMA datasets}
  \label{tab:prior-EMA-databases}
  \centering
  \eightpt
  \begin{tabular}{llllllll}
    \toprule
    \textbf{Corpus} &  \textbf{Language} & \textbf{Speakers} & \textbf{$f_s$ (Hz)} & \textbf{Sensors} & \textbf{\# Utterances/speaker} & \textbf{Conds} & \textbf{Availability} \\
    \midrule            
    \texttt{MOCHA-TIMIT} \cite{wrench1999-MOCHA-TIMIT} & ENG & 1\female, 1\male & 500 & 8 & 460 & L1 & Free \\
    \texttt{mngu0} \cite{richmond11_mngu0} & BrE & 1 & 200 & 6 & $\sim$1300 & L1 & Upon request \\
    \texttt{USC-TIMIT} \cite{Narayanan2014-realtime} & AmE & 2\female, 2\male & 100 & 6 & 460 & L1 & Upon request \\
    \texttt{EMA-MAE} \cite{Ji2014-EMA-MAE} & ENG & 10\female 10\male L1 ENG, & 400 & 7 & 45 mins & L1 ENG & Upon request \\
         &  & 10\female 10\male L1 CMN &  &  &  & L2 CMN accent &  \\
    \texttt{MSPKA} \cite{canevari_2015} & Italian & 1\male, 2\female & 400 & 7 & Varies; 500, 629, 666 & Conditions & Free \\
    \texttt{DKU-JNU-EMA} \cite{cai_dku-jnu-ema_2018} & 3 CMN dialects & 9\male, 7\female & NA & 6 & $\sim$0.66 hours & L1 dialect & Upon request \\
    \texttt{FROST-EMA} [this paper] & FIN, RUS & 8\female, 10\male & 1250 & 7 & L1 RUS 227, L1 FIN 223 & L1,L2,fake-L2 & Free\\
     
    \bottomrule
    \multicolumn{5}{l}{\footnotesize CMN refers to Mandarin Chinese. Only active sensors are included in the sensor column.} \\
  \end{tabular}
\end{table*}
\renewcommand{\arraystretch}{1.0}

\subsection{Materials}

In our study, native speakers of Finnish and Russian performed three speech production tasks: (1) reading a text paragraph, (2) reading a list of target words, and (3) performing an unconstrained picture description task under three different conditions (A, B and C). The three production tasks included
(1) reading the \textit{North Wind and the Sun} passage; (2) reading target words within a frame sentence---Finnish: \textit{Sano \_\_\_\_\_ uudestaan}, Russian: \textit{Povtori \_\_\_\_\_ snova} (``Say \_\_\_\_\_ again.'')---with 71 or 75 stimuli depending on their L1; and (3) an unconstrained picture description task \cite{gagarina2019main} assessing narrative abilities.

For Finnish native speakers, the design required them to (A) produce speech in their L1, (B) perform the same tasks while imitating a Russian accent, and (C) produce speech in Russian as their L2, naturally resulting in a Finnish accent. For native Russian speakers, the design followed the same structure: (A) speaking in their L1 (Russian), (B) imitating a Finnish accent while speaking Russian, and (C) completing the tasks in Finnish. This is a balanced design that allows for the study of both authentic and imitated accents.

The three production tasks and conditions give 
$3 \times 3 = 9$ types of data, recorded in the order: 1A, 1B, 1C, 2A, 2B, 2C, 3A, 3B, 3C. With the anticipated challenge related to occassional detachment of sensors from the speaker's tongue due to saliva secretion, the recording order reflects the authors' prioritization on language variation factors. 
The total number of recordings per participants are 227 for L1 Russian speakers and 223 for L1 Finnish speakers. The participants were recruited through internal channels of the University of Eastern Finland and were given two movie tickets as reward. Table~\ref{tab:participants} describes the participants in the dataset.

\begin{table}[b!]
  \caption{Participants in the data collection}
  \label{tab:participants}
  \centering
  \eightpt
  \begin{tabular}{llllll}
    \toprule
    \textbf{Participant ID}      & \textbf{Sex}   & \textbf{Age} & \textbf{L1} & \textbf{L2} & \textbf{L2 level} \\
    \midrule
    kh1 & M & 49 & RUS & FIN & average \\
    kh3 & M & 21 & FIN & RUS & basic \\
    kh4 & F & 39 & RUS & FIN & excellent \\
    kh5 & F & 21 & FIN & RUS & excellent \\
    kh6 & M & 21 & FIN & RUS & excellent \\
    kh7 & M & 35 & FIN & RUS & basic \\
    kh8 & F & 23 & FIN & RUS & good \\
    kh9 & M & 23 & FIN & RUS & average \\
    kh10 & F & 22 & FIN & RUS & good \\
    kh11 & M & 25 & FIN & RUS & good \\
    kh12 & F & 34 & FIN & RUS & excellent \\
    kh13 & F & 23 & FIN & RUS & excellent \\
    kh14 & F & 21 & FIN & RUS & excellent \\
    kh15 & F & 34 & RUS & FIN & beginner \\
    kh16 & M & 36 & RUS & FIN & excellent \\
    kh17 & M & 41 & RUS & FIN & beginner \\
    kh18 & M & 47 & RUS & FIN & excellent \\
    kh19 & M & 44 & RUS & FIN & good \\
    
    \bottomrule
  \end{tabular}
\end{table}

\subsection{Equipment}

The collected corpus consists of parallel acoustic (waveform) and articulatory (position) data.

\noindent\textbf{\underline{EMA:}} The state-of-the-art EMA device used was an AG501\footnote{\url{https://www.articulograph.de/articulograph-head-menue/the-ag501/} (URL valid as of \today).}, manufactured by Carstens Medizinelektronik GmbH. The position sampling frequency was 1250 Hz. 
The normalization process deducted the head movements of the participant from the data and left only the articulatory movements. 

\noindent\textbf{\underline{Audio:}} The number of audio channels was 1 (mono) and the sampling frequency was 48000 Hz. The microphone was a Sennheiser MKH 8060\footnote{\url{https://www.sennheiser.com/en-fi/catalog/products/microphones/mkh-8060/mkh-8060-506292} (URL valid as of \today)}, a shotgun microphone with high sensitivity (-24 dBV/Pa), based on RF (radio frequency) condenser transducer technology. The microphone was pointed downwards from the height of approximately 1.5 metres at the seated participant's forehead, 
positioned far away enough to 
suppress potential electromagnetic interference from the EMA device. The microphone was connected to 
Tascam US-2x2HR audio interface 
through a balanced XLR-3M cable. The recordings were carried out inside a Demvox DV676\footnote{\url{https://en.demvox.com/studio/dv-soundproof-booths/dv676} (URL valid as of \today).}, a soundproof booth designed for professional voice recordings.

\subsection{Data collection procedures and instructions}
\begin{figure}[t]
  \centering
  \includegraphics[width=0.8\linewidth]{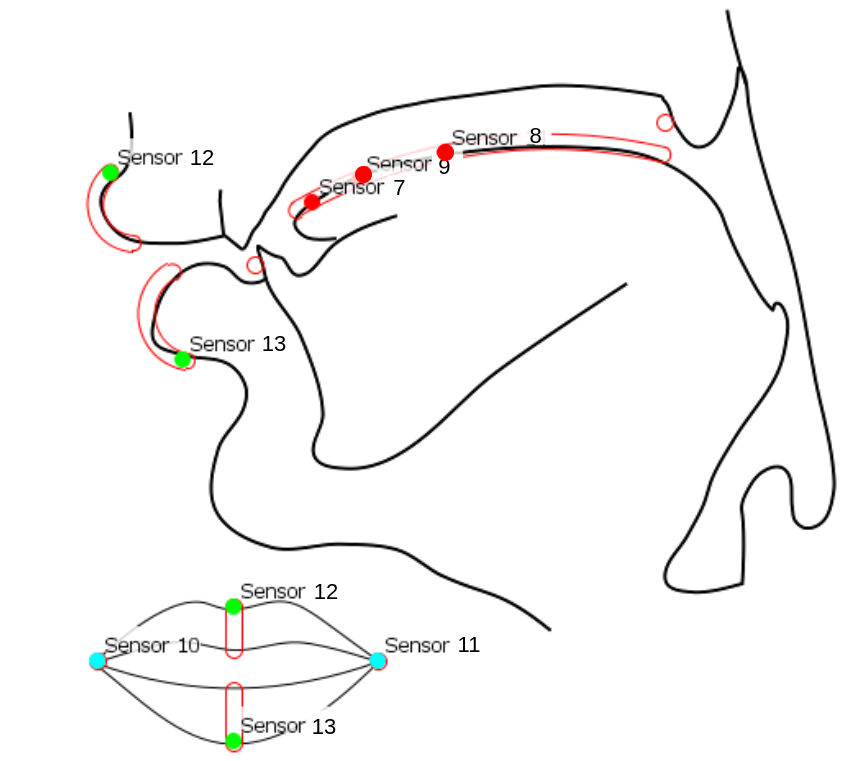}
  
  \caption{Sensor placement (visualized with \cite{ouni_visartico_2012}).}
  
  \label{fig:sensors_loc}
\end{figure}

The sensor protocol in our study was close to that of \cite{wang_optimal_2016} (for a review of different designs see \cite{rebernik_review_2021}), but with seven active sensors (see Fig.~\ref{fig:sensors_loc}). Four sensors were glued to the vermilion border of the lips, one in each corner and one at the midpoint of the top and bottom lip each. Three sensors were attached to the midline of the tongue one centimetre apart with the first a centimetre back from the tip. This ensured participant comfort and prevented the sensors from touching one another. We glued two reference sensors to the right and left mastoids, 
used for 
correcting for head movements in the normalization. 

Each recording session took between 45 and 75 minutes. Whenever the sensors needed regluing, this increased the time required.
The participants could request a break anytime, or report a detached sensor.

\subsection{Quality of the data}

While most subjects were recorded without issues, 
technical and hardware-related errors were encountered for two subjects (kh2---no audio recorded; kh3---pos data partially not recorded). We decided to keep data of kh3 but drop kh2. Additionally, some of the files have sensor errors in them. Some of the recording sessions did not exhibit such issues 
despite the same sensor sets being used, which leads us to believe that this a sporadic issue (see \cite{rebernik_review_2021} for similar remarks). The manufacturer's technical documentation did not provide any illumination.

EMA recordings face certain problems, which are expected and not insurmountable. As any electromagnetic device, EMA is liable to produce electrical interference. In our recordings, we noted 
high-frequency, low bandwidth, shimmering noise between 7000 and 13500 Hz. We therefore enhanced the audio using \textit{spectral de-noise} processing with iZotope RX 10 Advanced software. The adaptive mode settings included an amplitude separation threshold of 4 dB between noise and tonal signal levels, with a noise reduction level of 15.3 dB. This is not a problem for the acoustic analysis.

As is typical for EMA studies, the sensors and sensor wires cause the participants to lisp because the sensors sit on the surface of the tongue. 
For instance, kh3 (L1 Finnish) has difficulty pronouncing the voiced alveolar trill, [r], in Finnish and Russian both. He produces it consistently as approximant. 

\section{Case study I: Speaker verification}

\subsection{Setup}

We carried out an experiment to investigate the effects of conditions A, B, and C on automatic speaker verification (ASV). We used 1A as speaker enrollment. For testing, we used paired speech samples from stimuli section 2 with all variants (A, B and C). This allowed us to explore the performance of ASV with L1 as the baseline when presented with L2 speech and speech with an imitated accent. Table~\ref{tab:asvtrials} summarizes the number of speech pairs for the recognition experiment.

\begin{table}[htbp]
\caption{Description of the number of speaker trials for the recognition setup for female (8) and male (10) speakers.}
\centering
\eightpt
\begin{tabular}{l|rr|rr|r}
\cline{2-5}
\multicolumn{1}{l}{} & \multicolumn{2}{|c}{Female} & \multicolumn{2}{|c}{Male} & \multicolumn{1}{|l}{} \\ \cline{2-5}
\multicolumn{1}{l}{} & \multicolumn{1}{|l}{FIN} & \multicolumn{1}{|l}{RUS} & \multicolumn{1}{|l}{FIN} & \multicolumn{1}{|l}{RUS} & \multicolumn{1}{|l}{Total} \\ \hline
Same speaker & 1302 & 441 & 1085 & 1105 & 3933 \\ 
Different speaker & 9156 & 3045 & 9865 & 9845 & 31911 \\ \hline
\end{tabular}

\label{tab:asvtrials}
\end{table}

The ASV system we used is ECAPA-TDNN~\cite{desplanques20_interspeech}, a modern high-performance deep neural network model. We used the pre-trained model from the SpeechBrain toolkit~\cite{speechbrain} 
trained on the 16 kHz Voxceleb datasets~\cite{nagrani2020voxceleb}. 
The FROST-EMA data was downsampled accordingly. 
ECAPA-TDNN 
uses a time delay neural network (TDNN) model for speaker embedding extraction that incorporates squeeze-excitation layers, channel- and context-dependent statistics pooling, and multi-scale feature aggregation. The extracted speaker embeddings are 192-dimensional vectors.

\subsection{Results}
To explore the effect of the three conditions on the accuracy of ASV, we analyzed the performance reported in Table~\ref{tab:eer} in terms of equal error rate (EER). The EER for L1 speech (5.17\% for females, 3.33\% for males) is lower for the male speakers, which indicates that the ASV system performs better when speakers use their L1. When speakers produced L1 speech with an imitated L2 accent (condition B), the EER slightly increases (6.50\% for females, 3.60\% for males), which suggests that accent introduces variability, which, in turn, affects speaker verification. Interestingly, for L2 speech, the results diverge by gender: the EER for female speakers improves to 3.33\%, which matches the L1 result, but for male speakers, the EER increases to 4.27\%. This suggests that, at least in this study, gender-specific differences may influence ASV performance. 
All in all, our results indicate that although the ASV system remains effective across conditions, the use of a imitated accent and foreign language impacts performance. 

\begin{table}[htbp]
\caption{Performance of ASV in terms of equal error rate (\%)}
\centering
\eightpt
\begin{tabular}{lrr}
\toprule
\multicolumn{1}{l}{\textbf{Condition}} & \multicolumn{1}{l}{\textbf{Female}} & \multicolumn{1}{l}{\textbf{Male}} \\ \hline
A. L1 & 5.17 & 3.33  \\ 
B. L1 with imitated L2 accent & 6.50  & 3.60  \\ 
C. L2 & 3.33 & 4.27 \\ \hline
\end{tabular}

\label{tab:eer}
\end{table}

\begin{figure}[htbp]
\eightpt
  \centering
    \begin{subfigure}
    \centering
        \includegraphics[width=0.80\linewidth]{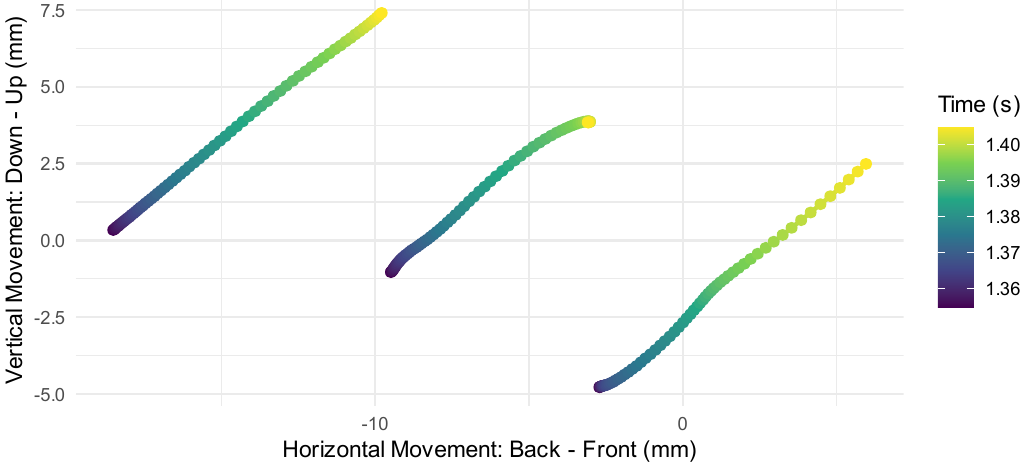}
        
        \caption{Tongue sensor trajectories (back, middle, tip) for a Finnish L1 speaker producing [æ] in the first syllable of \textit{hätä} (FIN).}
        \label{fig:fin_small}
    \end{subfigure}
\vspace{-5pt} 
    \begin{subfigure}
    \eightpt
    \centering
        \includegraphics[width=0.80\linewidth]{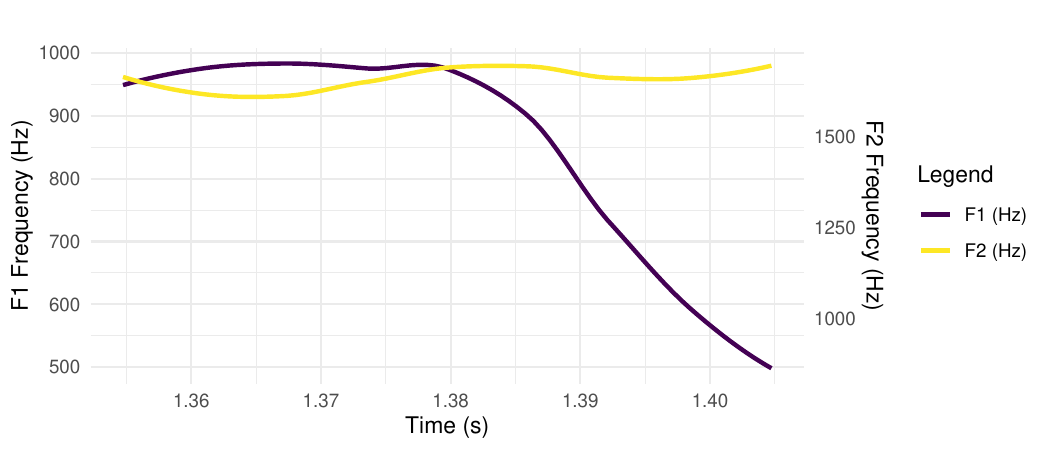}
        
        \caption{F1 and F2 frequencies over time; Finnish L1 speaker producing [æ] in the first syllable of \textit{hätä} (FIN).}
        \label{fig:fin_f1f2}
    \end{subfigure}
\end{figure}

\begin{figure}[htbp]
\eightpt
  \centering
    \begin{subfigure}
    \centering
        \includegraphics[width=0.95\linewidth]{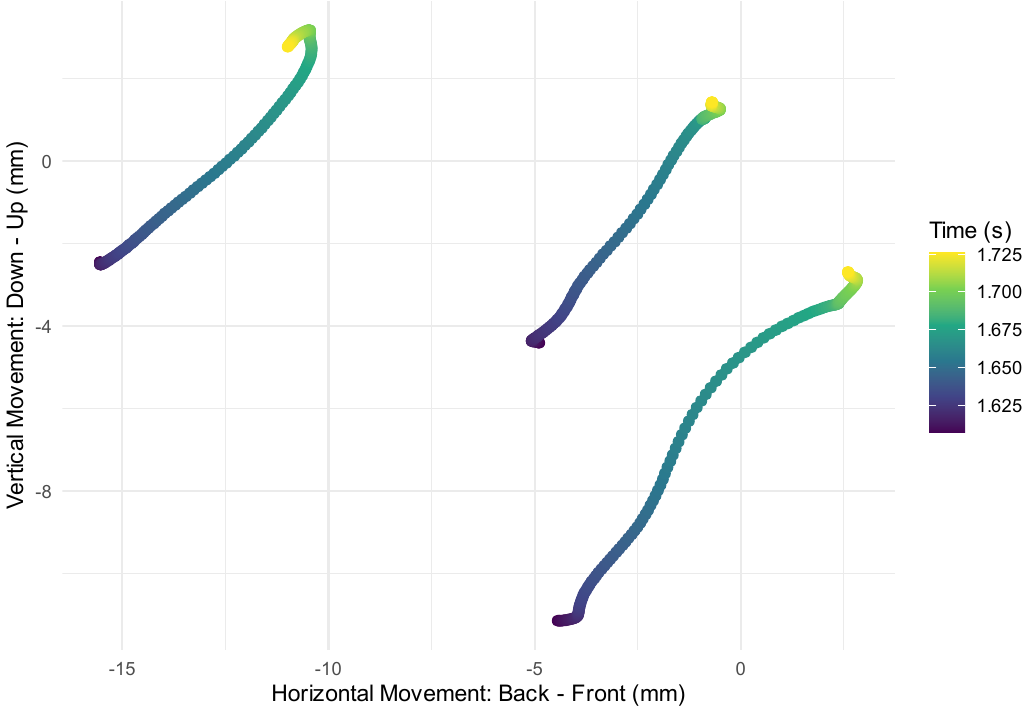}
        
        \caption{Tongue sensor trajectories (back, middle, tip) for a Finnish L1 speaker producing [æ] in the stressed first syllable of \textit{hätä} with imitated L2 accent.}
        \label{fig:fake_fin_small}
    \end{subfigure}
   \vspace{-5pt} 
    \begin{subfigure}
    \eightpt
        \centering
        \includegraphics[width=0.95\linewidth]{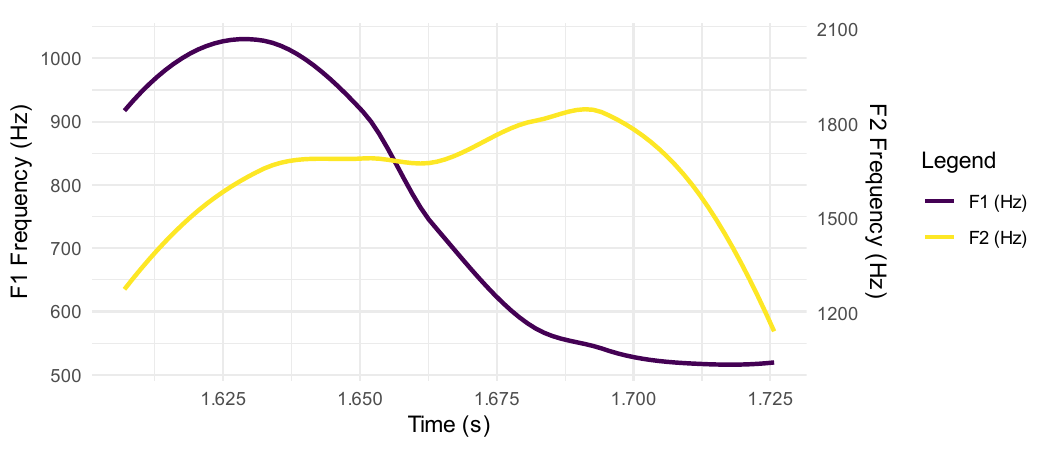}
        
        \caption{F1 and F2 frequencies for a Finnish L1 speaker producing [æ] in the stressed syllable of \textit{hätä} with imitated L2 accent.}
        \label{fig:fake_fin_f1f2}
    \end{subfigure}
 \end{figure}

\begin{figure}[htbp]
\eightpt
  \centering
    \begin{subfigure} 
        \centering
        \includegraphics[width=0.95\linewidth]{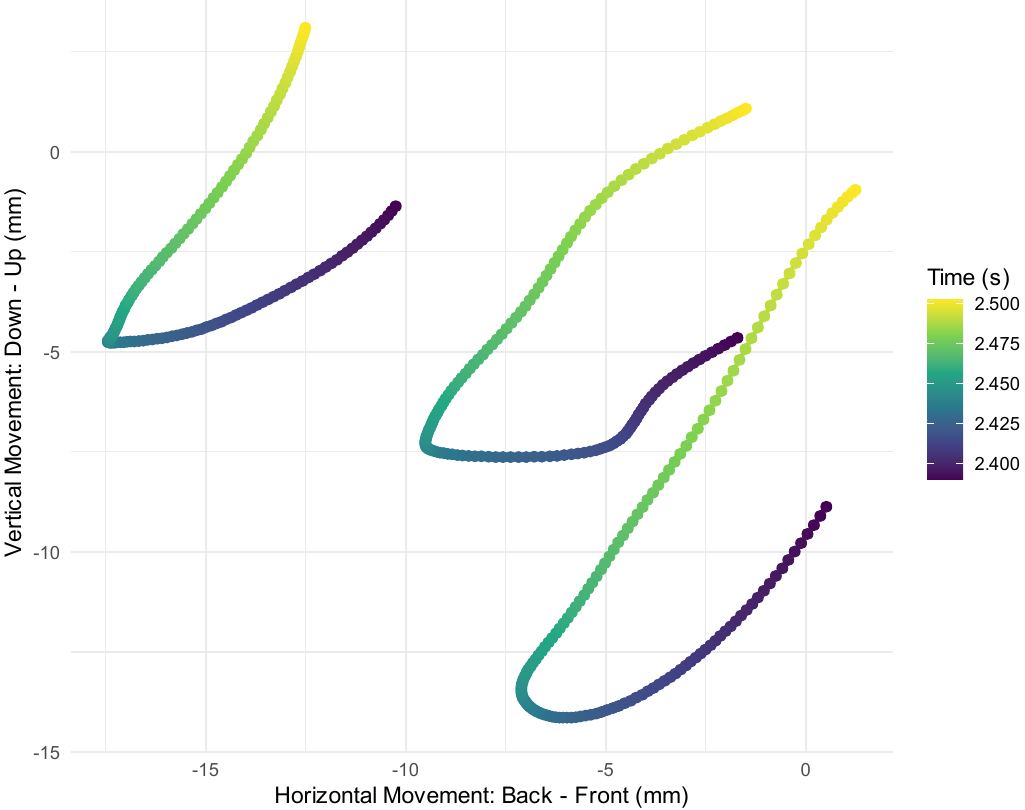}
        \caption{Tongue sensor trajectories (back, middle, tip) for a Finnish L1 speaker producing [æ] in /\textit{p\textsuperscript{j}æt\textsuperscript{j}}/ (L2, RUS).}
        \label{fig:rus_small}
    \end{subfigure}
  \vspace{-5pt} %
    \begin{subfigure} 
        \eightpt
        \centering
        \includegraphics[width=0.95\linewidth]{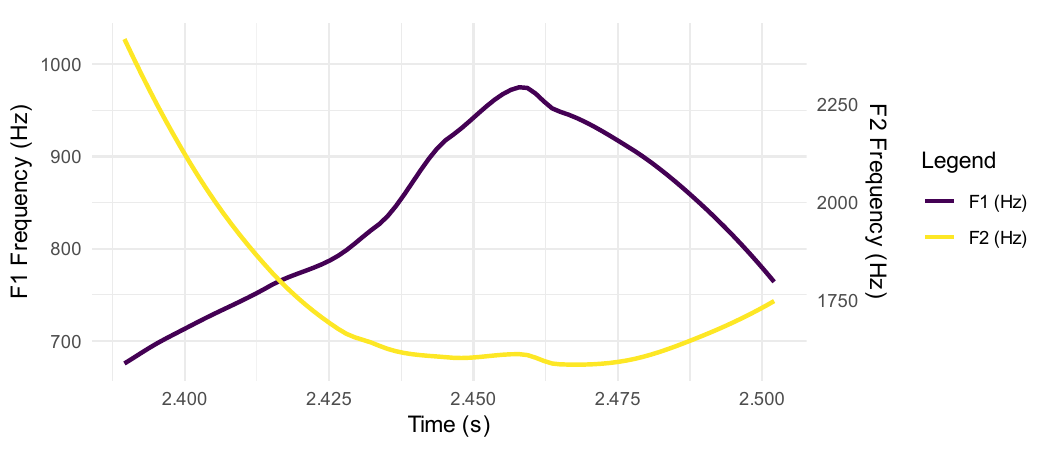}
        \caption{F1 and F2 frequencies over time; Finnish L1 speaker producing [æ] in /\textit{p\textsuperscript{j}æt\textsuperscript{j}}/ (L2, RUS).}
        \label{fig:rus_f1f2}
    \end{subfigure}
\end{figure}

\section{Case Study II: 
Articulation 
of [æ]} 

kh5 (a female participant with L1 Finnish and L2 Russian) provides an illustrative example of the insights that EMA data can offer. We extracted formant measurements with \cite{praat}. In the figures below, co-articulation effects caused by the surrounding consonants can be seen in the movements of the three tongue sensors (Figures \ref{fig:fin_small}, \ref{fig:fake_fin_small}, \ref{fig:rus_small}.) In addition, the articulatory trajectories of three [æ] vowels, which are expected to exhibit similar movement patterns because of their acoustic similarity (Figures \ref{fig:fin_f1f2}, \ref{fig:fake_fin_f1f2}, \ref{fig:rus_f1f2}) demonstrate overlapping gestures. Further, under the L2 condition (Russian), a peculiar backward–forward movement pattern of the tongue is evident (Fig. \ref{fig:rus_small}), suggesting that [æ] is produced either as a diphthong or with a preceding semivowel [j] (which is also evident in F1 and F2 (Fig. \ref{fig:rus_f1f2}). This suggests that  a non-imitated foreign accent may induce articulatory adjustments that differ from the typical patterns observed in L1 speech and from a fake foreign accent.

\vspace{-5pt} 

\vspace{-10pt} 

These visualizations underscore the novelty of EMA data. The plots reveal dynamic articulatory processes that remain unobserved from mere audio recordings. Although our current analysis of subject kh5 is for illustration purposes only and does not allow for statistical analysis, the abundance of our EMA data allows for detailed statistical analyses in future work.


It is also important to note that, to our knowledge, no EMA database provides a two-directional investigation of accent production (including both native and imitated foreign accent conditions) apart from the small dataset collected by \cite{tavi_2021}. This opens new avenues for studying the articulatory correlates of accent variation.

Taken together, these findings illustrate both the potential and the challenges of analyzing co-articulation and accent effects using EMA data. Future studies may build on our preliminary analysis to quantify these articulatory differences.

\section{Discussion}

Our new FROST-EMA corpus with an innovative design introduced in this work is designed to serve purposes of both speech science and speech technology for bilingual analysis. Following metadata clean-up and further auditing of the individual sentences, we anticipate the release of the database in the first half of 2026.


\section{Acknowledgements}
The authors wish to thank the School of Computing of the University of Eastern Finland and the Logolab research environment located at the School of Humanities, University of Eastern Finland, for the use of its facilities. In addition, the authors wish to thank senior laboratory technician Petri Pulli and research assistants Martta Sinettä, Annisa Wardhani and Jaka Wibawa. This work was partially supported by Academy of Finland (Decision No. 349605, project ”SPEECHFAKES”).


\bibliographystyle{IEEEtran}
\bibliography{mybib}

\end{document}